\documentclass[letterpaper, 10 pt, journal, twoside]{IEEEtran}  
\usepackage{amsmath}
\usepackage{amssymb}
\usepackage{graphicx}
\usepackage{cite}
\usepackage{hyperref}
\usepackage{booktabs}
\usepackage{multirow}
\usepackage{microtype}
\usepackage{algorithm}
\usepackage{algorithmic}
\usepackage{svg}
\usepackage{amsmath}
\usepackage{tikz}
\usepackage{graphicx}
\usepackage{float}
\usepackage{diagbox}
\usepackage{mathptmx}
\usepackage{authblk}

\usetikzlibrary{bayesnet} 

\usetikzlibrary{shapes,arrows,positioning,calc}

\providecommand{\mathbf}[1]{\boldsymbol{#1}}
\providecommand{\mathrm}[1]{\textrm{#1}}

\svgsetup{
    inkscapelatex=false,
    svgpath=figures/   
}

\IEEEoverridecommandlockouts                              

\title{\LARGE \bf
Empowering a Single-Frequency GNSS Receiver to Achieve High-Precision Positioning with Relative Observations
}

\author{Xingpeng Wang$^{1,2}$, Ziwen Qu$^{1,2}$, Juncheng Chen$^{1,2}$, Ruitian Pang$^{1,2}$, Xiangyu Li$^{1,2}$, Tiancheng Lai$^{1,2}$, \\ Siqi Shen$^{1,2}$, Wentao Liu$^{1,2}$, Pengfei Wang$^{2}$, Chao Xu$^{1,2}$, and Yanjun Cao$^{1,2,*}$ 
\thanks{$^{1}$State Key Laboratory of Industrial Control Technology, Institute of Cyber Systems and Control, Zhejiang University, Hangzhou, China.}
\thanks{$^{2}$Huzhou Institute of Zhejiang University, Huzhou, China.}
\thanks{$^{*}$Corresponding author. This work was supported by the Key R\&D Project of China National Tobacco Corporation under Grant No. 110202402018.}}

\begin{document}

\maketitle
\thispagestyle{empty}
\pagestyle{empty}

\begin{abstract}
	
	Global Navigation Satellite System (GNSS) navigation is widely used to provide absolute, outdoor positioning in field robotics. 
	Advances in Real-Time Kinematic (RTK) technology can achieve centimeter-level accuracy, facilitating autonomous navigation tasks.
	However, the cost and extra infrastructure used for RTK still hinder the application and more cost-effective solutions are desired.
	In this letter, we present a novel tightly-coupled state estimation framework that achieves high-precision localization by using low-cost, mass-market single-frequency GNSS receivers with any relative motion sensors (e.g., wheel encoder, camera, LiDAR). We propose a sliding-window factor graph that integrates generic relative motion with global epoch-to-anchor constraints derived from continuous carrier phase tracking. 
	To eliminate the reliance on physical base stations, we introduce a \textit{virtual anchor} mechanism: upon the initial observation of a satellite, its state is locked as a virtual reference to establish global epoch-to-anchor constraints. 
	By substituting multi-frequency hardware redundancy with single-frequency multi-modal kinematic priors and a robust cycle-slip recovery technique, our approach ensures carrier-phase integrity on cheap receivers. 
	Extensive real-world experiments on heterogeneous low-cost sensor suites validate that our method improves the accuracy of a single-frequency receiver from several meters to decimeter-level precision across diverse environments, providing an accurate, cost-effective and reliable alternative for autonomous navigation.
	
\end{abstract}

\section{Introduction}
\IEEEPARstart{A}{ccurate} localization is a fundamental prerequisite for autonomous mobile robots operating in complex outdoor environments~\cite{stempfhuber2012precise, zhao2013quaternion}, with Global Navigation Satellite Systems (GNSS) serving as the primary source for absolute positioning. 
Real-Time Kinematic (RTK)~\cite{moon2016outdoor} technology is frequently employed to achieve centimeter-level precision. 
However, the usage of RTK relies heavily on continuous, reliable communication with base-station infrastructure and incurs high hardware costs, which limits its application in independent or large-scale missions. 
Similarly, standalone multi-frequency~\cite{shengPrecisePointPositioning2020} receivers, although realize decimeter-level accuracy while mitigating the need for base stations, still suffer from extended convergence time and extra financial overhead.
Consequently, there is an highly desired 
need to develop robust, cost-effective solutions capable of delivering high-precision localization without relying on expensive hardware or external reference networks. Single-frequency GNSS receivers, which are widely used in consumer electronics and low-cost robotics, present a promising alternative due to their affordability and ubiquity.

To achieve high-precision localization performance with low-cost single-frequency GNSS receivers is not trivial. 
Current state-of-the-art methods still face the following challenges:
\begin{itemize}
	\item \textbf{Noisy pseudorange} measurements to provide absolute positioning, but their inherent noise generally limits the achievable accuracy to the meter level.
	\item \textbf{Drifted time difference} carrier phase~\cite{fredaTimedifferencedCarrierPhases2015} between adjacent epochs (TDCP) provides highly accurate relative position constraints if cycle slip is properly handled, but inherently suffers from drift over long trajectories.
	\item \textbf{Misaligned cycle slips} in carrier phase cause abrupt changes in the integer ambiguity, leading to significant errors if not properly detected and corrected. 
\end{itemize}

In this paper, we propose a novel tightly-coupled state estimation framework combining relative motion sensor (e.g., LiDAR, camera, or wheel encoder) to tackles the above challenges and achieve high-precision localization.
Our system accepts generic relative observations as short-term kinematic priors to detect possible cycle slips, ensuring the integrity of carrier phase measurements.
Building upon this phase tracking, we introduce epoch-to-anchor constraint calculating the current robot state via an initializatied anchor. 
By enforcing these long-term, absolute geometric constraints within a sliding-window factor graph, our system realizes high accuracy. 
We extensively validate the proposed framework across diverse outdoor scenarios and multiple sensor configurations. 
Experimental results demonstrate that our method consistently achieves a translation root-mean-square error (RMSE) of decimeter-level, significantly outperforming state-of-the-art algorithms for a single-frequency GNSS receiver. 
The main contributions of this paper are summarized as follows:

\begin{figure*}[ht!]
    \centering
    \includegraphics[width=0.9\textwidth]{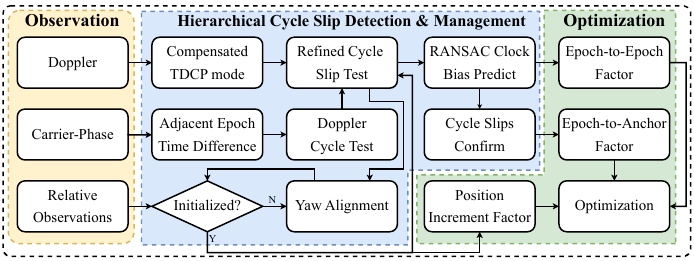}
    \caption[System architecture of the proposed localization framework.]{
        \textbf{System architecture of the proposed localization framework.} 
        The framework comprises three primary modules: 
        (1) \textbf{Observation}, which synchronizes raw GNSS observations (doppler and carrier-phase) with generic relative observations; 
        (2) \textbf{Hierarchical Cycle Slip Detection and management}, which leverages doppler integration and relative motion priors to robustly detect and recover cycle slips; and 
        (3) \textbf{Optimization}, which fuses the local registration factor, the adjacent TDCP factor, and the epoch-to-anchor factor to achieve low-drift and accurate state estimation.
    }
    \label{fig:system_overview}
\end{figure*}                                                                                                                                                                                                                  

\begin{itemize}
	\item \textbf{Novel factor graph formulation} introduces a virtual anchor carrier phase difference     (epoch-to-anchor) constraint which significantly mitigates the cumulative error.
	\item \textbf{Coarse-to-fine initialization} ensures reliable calibration between GNSS and local sensors even in the presence of cycle slips during start-up.  
	\item \textbf{Ambiguity monitor} utilizes relative position increments in phase changes to detect slips and recover integer ambiguities for global ambiguity states.
	\item \textbf{Random walk clock bias estimation} ensures robustness when the number of satellites is insufficient for clock bias estimation, which is common in low signal conditions.
\end{itemize}

\section{Related Work}

\subsection{Single-frequency Cycle Slip Handling Algorithms}
The primary obstacle to extracting global constraints from single-frequency carrier phase measurements is the integer ambiguity, which changes unpredictably upon cycle slips caused by signal obstruction or high dynamics. Double-differenced carrier-phase measurement has been used as the test statistic~\cite{khan2010locata}, but it requires four consecutive epoch measurements, resulting in a time delay. Doppler residuals have also been used for cycle slip detection~\cite{liSinglefrequencyCycleSlip2022}, but it suffers from limited effectiveness in dynamic environments and is unavailable in low signal conditions such as low number of valid satellites and poor signal-to-noise ratio. Fully odometry is also tried to assist ambiguity resolution~\cite{xuSinglestationSinglefrequencyGNSS2024}. However, relying on a full SLAM system for cycle slip detection imposes a severe computational burden. Furthermore, it ignores potential cycle slips during the frame alignment phase, as well as clock bias drift between adjacent epochs. While Inertial Navigation System aided cycle slip detection has also been proposed~\cite{kimGPSCycleSlip2015}~\cite{shenNovelFactorGraph2024}, its reliance on high-accuracy inertial sensors incurs unaffordable hardware costs.
\subsection{Single-frequency GNSS Localization Algorithms}
While pseudorange-based single point positioning (SPP)~\cite{cao2022gvins} provides absolute localization, its inherent noise usually limits accuracy to the meter level. TDCP~\cite{fredaTimedifferencedCarrierPhases2015} measurement can provide highly accurate relative constraints when cycle slips are properly handled, but integrating only relative constraints still causes unbounded drift over long trajectories~\cite{zhaoHighPrecisionVehicleNavigation2016}. To mitigate this issue, epoch-to-anchor TDCP~\cite{zhaoApplyingTimeDifferencedCarrier2017,baiPerformanceEnhancementTightly2024} directly constrains the current state to a static initialization anchor. However, this formulation is still highly sensitive to cycle slips, which introduce abrupt ambiguity changes and large positioning errors if not detected and corrected. A combined epoch-to-anchor TDCP and cycle-slip detection strategy was proposed in~\cite{liuOptimizationBasedVisualInertialSLAM2021}, but it assumes dual-frequency observations for slip detection and resets anchors after slips, which is unfavorable for long-duration missions. A momentary loss of lock on one frequency can therefore invalidate its correction mechanism. Ambiguity-free methods~\cite{yangImprovedRelativeGNSS2020}~\cite{hedgecockAccurateRealtimeRelative2014} have also been proposed, but they typically mandate a cooperative network of multiple receivers and relies heavily on continuous inter-node data transmission.
\subsection{GNSS-sensors Fusion Localization Algorithms}
To achieve absolute positioning, tightly coupled SLAM systems frequently fuse GNSS pseudorange and doppler observations\cite{hanTightlyCoupledOptimizationbased2022, cioffiTightlycoupledFusionGlobal2020, liu2023glio, liP3LINSTightlyCoupled2023}, since these measurements are free of cycle slips, they provide robust, albeit noisy, global constraints. Notably, LIGO\cite{he2025ligo} goes a step further by utilizing epoch-to-epoch carrier phase measurements to constrain the local trajectory. However, it assumes ideal phase tracking and lacks an explicit mechanism for cycle slip detection and recovery, rendering it vulnerable in highly dynamic or occluded environments. Furthermore, exsiting GNSS-sensor fusion solutions are predominantly sensor-specific. Their strict reliance on maintaining dense global or local submaps inevitably incurs prohibitive computational overhead and hardware costs, rendering them unsuitable for low-cost, long-endurance missions. 

\section{Problem Statement}
\subsection{Coordinate Frames}
\begin{figure}[htbp]
	\centering
	\includegraphics[width=1.0\linewidth]{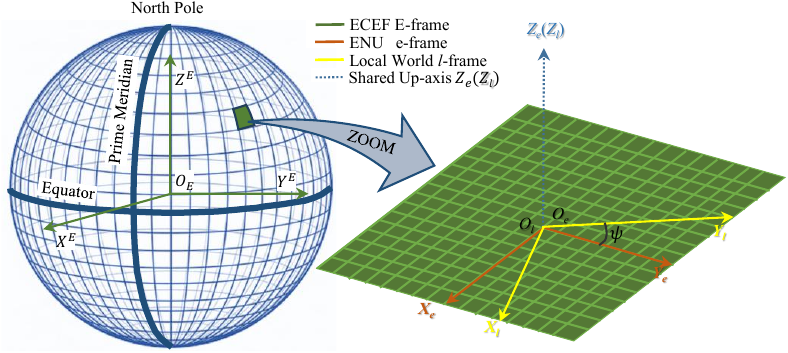}
	\caption{Coordinate frames used in the proposed system.}
	\label{fig:frame}
\end{figure}

To facilitate the fusion of global GNSS measurements with local sensor measurements, we define three coordinate frames as illustrated in Fig.~\ref{fig:frame}:

\begin{itemize}
    \item \textbf{ECEF Frame ($\mathcal{F}_E$):} The Earth-Centered, Earth-Fixed global frame. Its origin is the Earth's center of mass, with the X-axis pointing to the intersection of the equator and the prime meridian, the Z-axis pointing to the North Pole, and the Y-axis completing a right-handed orthogonal system.
    \item \textbf{ENU Frame ($\mathcal{F}_{e}$):} The local geographic frame anchored at the initial position $^E\mathbf{p}_{\text{anc}}$. Its X, Y, and Z axes align with the local East, North, and Up directions, respectively.
    \item \textbf{Local Frame ($\mathcal{F}_l$):} The gravity-aligned local odometry frame. Since its Z-axis coincides with the Up direction of $\mathcal{F}_{e}$, the spatial transformation between $\mathcal{F}_l$ and $\mathcal{F}_{e}$ is simplified to a single yaw angle $\psi$.
\end{itemize}

The transformation chain from a point in the local world frame $\mathbf{p}^l$ to the global ECEF frame $^E\mathbf{p}$ is formulated as:
\begin{equation}
    ^E\mathbf{p} = \mathbf{p}_{\mathrm{\text{anc}}} + {}^{E}\mathbf{R}_{\mathrm{e}}{}^{\mathrm{e}}\mathbf{R}_{l}\left(\psi\right){}^{l}\mathbf{p}
\end{equation}
Here, ${}^{E}\mathbf{R}_{\mathrm{e}}{}$ is the rotation matrix derived from the anchor's geographic coordinates, and ${}^{\mathrm{e}}\mathbf{R}_{l}(\psi)$ represents the rotation around the Z-axis aligning $\mathcal{F}_l$ with $\mathcal{F}_{e}$.


\subsection{State and Problem Formulation}
To formulate the problem, we first define the system state and the available observations. The state vector of our system at time step $k$ in $\mathcal{F}_{e}$ is defined as:
\begin{equation}
    \mathbf{x}_k^e \triangleq \{ {}^e\mathbf{p}_k, \Delta \delta {t}_k, \psi_k \},
\end{equation}
where ${}^e\mathbf{p}_k$ is the position of the receiver in the ENU frame. $\Delta \delta {t}_k$ is the receiver clock bias variations from the initial epoch, and $\psi_k$ denotes the yaw angle aligning the local frame to the global ENU frame. 
The full trajectory of states up to time step $k$ is denoted as $\mathbf{\chi}_k \triangleq \{ \mathbf{x}_0, \ldots , \mathbf{x}_k \}$.

\textbf{Observations and System Parameters:} 
Our system utilizes relative motion $\mathcal{Z}_{k}^{\text{rel}} = \{ \Delta^l \mathbf{p}_k \}$ from generic local sensors alongside raw single-frequency GNSS carrier phase and doppler measurements $\mathcal{Z}_{k}^{\text{GNSS}} = \{ \Phi_k, \mathcal{D}_k \}$. Crucially, we define the incremental integer ambiguities $\Delta {\mathbf{N}}_k$ as \textit{explicit parameters} for each satellite which are pre-determined by our cycle slip detection frontend in \ref{sec:cycleslipPar}. 

\textbf{Relative and GNSS Observations:} 
Our system relies on two continuous streams of measurements. The first is the relative motion observations $\mathcal{Z}_{k}^{\text{rel}} = \{ \Delta^l \mathbf{p}_k \}$ derived from the generic local sensor suite between consecutive steps. The second is the raw GNSS measurements $\mathcal{Z}_{k}^{\text{GNSS}} = \{ \Phi_k, \mathcal{D}_k \}$, consisting of carrier phase and doppler observations.

\textbf{Problem Formulation:} 
To address the vulnerability of single-frequency carrier phases to cycle slips without relying on multi-frequency hardware, we utilize $\mathcal{Z}_{k}^{\text{rel}}$ not merely as local odometry, but as a strict kinematic prior serving the essential criterion to manage cycle slips.

With phase continuity safeguarded by our cycle slip test mechanism, the optimal trajectory $\mathbf{\chi}_k$ is estimated via Maximum A Posteriori (MAP) inference. Assuming Gaussian noise, we minimize the squared Mahalanobis distances over a sliding-window factor graph:
\begin{align}
    \hat{\mathbf{\chi}}_k &= \arg\max_{\mathbf{\chi}_k} P(\mathbf{\chi}_k \mid \mathcal{Z}_{1:k}^{\text{rel}}, \mathcal{Z}_{1:k}^{\text{GNSS}}) \notag \\
    &= \arg\min_{\mathbf{\chi}_k} \left( \sum_{k} \left\| \mathbf{r}_{\text{rel}}(\mathbf{x}_{k-1}, \mathbf{x}_k, \mathcal{Z}_{k}^{\text{rel}}) \right\|_{\Sigma_{\text{rel}}}^2 \right. \notag \\
    &\quad \left. + \sum_{k} \left\| \mathbf{r}_{\text{GNSS}}(\mathbf{x}_k, \mathcal{Z}_{k}^{\text{GNSS}} | \mathcal{Z}_{k}^{\text{rel}}) \right\|_{\Sigma_{\text{GNSS}}}^2 \right),
\end{align}
where $\mathbf{r}_{\text{rel}}(\cdot)$ bounds the local trajectory, acting as the continuous prior that safeguards the global epoch-to-anchor GNSS phase residual $\mathbf{r}_{\text{GNSS}}(\cdot)$ against unmodeled slips. $\Sigma_{\text{rel}}$ and $\Sigma_{\text{GNSS}}$ are their respective measurement covariance matrices.

\section{Methodology}
Building upon the state definitions and coordinate frames established above, this section elaborates on the proposed localization pipeline. We first introduce the hierarchical cycle slip detection and recovery mechanism, which is critical for maintaining the integrity of carrier phase measurements. Then, we detail the coarse-to-fine frame alignment process. Finally, we present the joint factor graph optimization that fuses local position increments, adjacent carrier phase differences, global epoch-to-anchor constraints and random walk clock bias estimation to achieve low-drift localization. The overall system architecture is illustrated in Fig.~\ref{fig:system_overview}.

\subsection{Accurate Cycle Tracking with Relative Observations}\label{sec:cycleslipPar}
The carrier phase observation $\Phi_{k}^j$ from satellite $j$ to receiver $r$ at time step $k$ provides a precise distance measurement, formulated as:
\begin{equation}
    \lambda \Phi_k^j = \rho_k^j + c(\delta{{t}_k} - \Delta{{t}_k^j}) + \lambda N_k^j + I_k^j + T_k^j + \epsilon_{k,\Phi}^j ,
\end{equation}
where $\lambda$ is the signal wavelength and $\rho_k^j = \left\| ^E\mathbf{p}_k^j - ^E\mathbf{p}_k \right\|_2$ is the geometric distance linking the receiver position $\mathbf{p}_k$ and satellite position $\mathbf{p}_k^j$. The terms $\delta t_k$ and $\Delta t_k^j$ (derived from ephemeris) represent receiver and satellite clock biases, while $c$ is the speed of light. $N_k^j$ denotes the integer ambiguity. The ionospheric $I_k^j$ and tropospheric $T_k^j$ delays are compensated using the Klobuchar \cite{klobucharIonosphericTimeDelayAlgorithm1987} and Saastamoinen \cite{saastamoinenContributionsTheoryAtmospheric1972} models, respectively, and $\epsilon_{k,\Phi}^j$ accounts for measurement noise.


While carrier phase observations offer superior precision, resolving their unknown integer ambiguities remains challenging. Thus, TDCP is widely adopted to eliminate the constant ambiguity via temporal differencing. The TDCP measurement $\Delta\Phi_{{k}}^{j}$ for satellite $j$ between consecutive steps $k-1$ and $k$ is formulated as:
\begin{align}
	\lambda\Delta\Phi_{{k}}^{j} &= \rho_{k}^j - \rho_{k-1}^j + c({\delta t_{k} - \delta t_{k-1}}) - c(\Delta t_k^j - \Delta t_{k-1}^j) \notag \\
	&\quad + \lambda({N}_k^j - {N}_{k-1}^{j}) + \sigma_{{k,\Phi}}^j.
\end{align}

\subsubsection{Relative Observation Priority}

We predict the receiver position $^e\mathbf{\hat p}_k$ at time step $k$ using the relative local motion $\Delta^l\mathbf{\hat p}_k$ and the prior state $^e\mathbf{\check p}_{k-1}$:
\begin{equation}
    ^e\mathbf{\hat p}_k = {}^e\mathbf{\check p}_{k-1} + {}^e\mathbf{R}_l(\check \psi_{k-1}) \Delta^l\mathbf{\hat p}_k.
\end{equation}

To achieve robust cycle slip detection and clock bias estimation, we propose a hierarchical scheme. First, we use the ambiguity-free mean integrated Doppler $\Delta \Phi_{\mathcal D,k}^j \approx \frac{\mathcal D^j_{k-1} + \mathcal D^j_k}{2} \Delta t$ to predict phase changes. Satellites whose Doppler-phase residual $\gamma_{\mathcal D, k}^j = \Delta \Phi_{k}^j - \Delta \Phi_{\mathcal D,k}^j$ exceeds a coarse threshold $\tau_{c}$ are excluded, yielding a high-confidence subset $\mathcal{S}_k$.

\subsubsection{Refine Clock Bias Estimation with RANSAC}
Next, we refine the receiver clock bias change $c \Delta \delta t_k$ by applying a RANSAC scheme on $\mathcal{S}_k$. We formulate the individual phase residual $\gamma_{\Phi,k}^j$ based on the predicted geometric distance change:
\begin{equation}
    \gamma_{\Phi,k}^j = \Delta \Phi_{k}^j - (\hat{\rho}_k^j - \check \rho_{k-1}^j),
    \label{eq:cycle_test_residual}
\end{equation}
where $\hat{\rho}_k^j = \| ^E\mathbf{p}_k^j - (^E\mathbf{p}_{\text{anc}} + {}^E\mathbf{R}_l(\check \psi_{k-1}) \Delta^l\mathbf{\hat p}_k) \|$. 
Using a strict threshold $\tau_s$, RANSAC iteratively finds the maximum inlier set $\mathcal{I}^*_k \subset \mathcal{S}_k$. To ensure robustness against remaining outliers, the optimal clock bias change is then determined as the median of these inlier residuals: $c \Delta{\delta \hat t_{k}} = \operatorname*{median}_{j \in \mathcal{I}^*_k} ( \gamma_{\Phi,k}^j )$.

\begin{algorithm}[htbp]
\caption{Robust Cycle Slip Detection \& Status Update}
\label{alg:cycle_slip_process}
\small 
\begin{algorithmic}[1]
\REQUIRE Observations $\mathcal{O}_k$, Relative Motion $ \Delta \mathbf{p}_{k} ^{l} $, Cache $\mathcal{C}$
\ENSURE Updated Cache $\mathcal{C}$

\STATE \textbf{Stage 1: Reference Subset Selection}
\STATE Define candidate set $\mathcal{S}^{*}$ using doppler consistency:
\STATE $\mathcal{S}^{*} \leftarrow \{ j \in \mathcal{O}_k \mid \text{SNR}^j > \tau_{high} \land |\Delta \Phi_k^j - \Delta \hat{\Phi}_{\mathcal{D}}^j| < \tau_{c} \}$

\STATE \textbf{Stage 2: Robust Clock Estimation}
    \STATE $c \Delta \hat{\delta} t_k \leftarrow \text{RANSAC}(\mathcal{S}^{*}, \Delta  \mathbf{p}_{k}^{l}, \Delta {\Phi}_{k}^{j})$
\STATE \textbf{Stage 3: Global Verification \& Branching}
\FOR{each satellite $j$ in $\mathcal{O}_k$}
    \STATE Calculate $r^j_k \leftarrow |\lambda\Delta \Phi_k^j - \Delta \rho_{k}^j(\Delta  \mathbf{p}_{k}^{l}) - c\Delta\hat{\delta} t_k|$
    \IF{$r^j_k > \tau_{s}$}
        \STATE $\mathcal{C}.\text{SetStatus}(j, \texttt{Drop})$; \textbf{continue}
    \ENDIF
    
	\IF{$\mathcal{C}.\text{IsHold}(j)$ \OR ($\mathcal{C}.\text{IsStable}(j) \land \sum_{m=0}^{M-1} \mathbb{I}(|\gamma_{k-m}^j| < \tau_{s}) = M $)}
		\STATE $\Delta \check N_{k}^j$ $\leftarrow$ $\Delta \check N_{k-1}^j + \Delta \check N_{k-1,k}^{j}$
		\STATE $\mathcal{C}.\text{SetStatus}(j, \texttt{Hold}, \Delta \check N_{k}^j)$ 
	\ENDIF
\ENDFOR

\RETURN $\mathcal{C}$

\end{algorithmic}
\label{alg}
\end{algorithm}
Exploiting the integer nature of carrier phase measurements, we detect cycle slips by thresholding the phase residual (Alg.~\ref{alg}). An observation is flagged as a cycle slip if $| r(\mathbf{\hat x}_k, \Delta{\Phi^{j}_k}) | > \tau_s$. The incremental integer ambiguity change $\Delta \hat{N}_{{k}}^{j}$ is then recovered by rounding the residual to the nearest integer:
\begin{equation}
    \Delta \hat{N}_{{k-1, k}}^{j} = \text{round} \left( \frac{r(\mathbf{\hat x}_k, \Delta{\Phi^{j}_k})}{\lambda} \right).
\end{equation}

\subsubsection{Ambiguity Management} 
To maintain the reliability of the factor graph, we propose a hysteresis-based satellite status management strategy comprising two states: \texttt{Hold} and \texttt{Drop}.

\begin{figure*}[t!]
    \centering
    \includegraphics[width=0.9\linewidth]{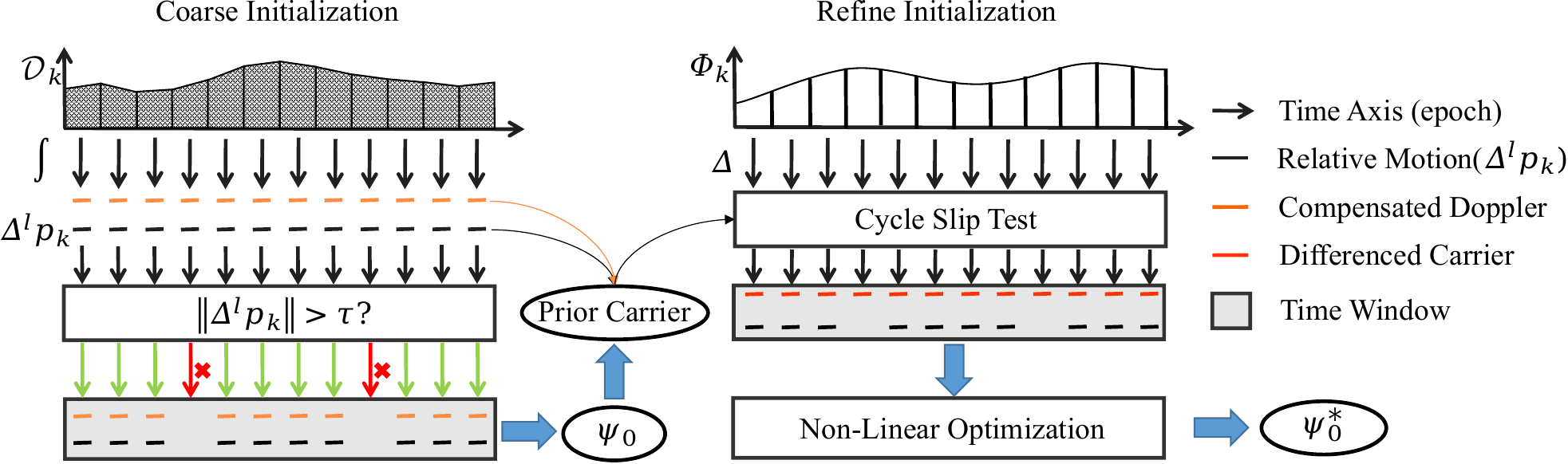} 
    \caption{\textbf{Initialization framework.} The system first performs a coarse alignment using Doppler integration to estimate the initial yaw angle and clock bias. Subsequently, it refines the frame alignment using TDCP measurements, guided by the initial coarse estimates.}
    \label{fig:initialization_framework}
\end{figure*}

\noindent \textbf{Status Maintenance:} 
When a new satellite is observed, it is initially flagged \texttt{Hold} as a virtual anchor of the satellite. For a satellite already flagged \texttt{Hold}, if its residual in the adjacent epoch satisfies $|\gamma_k^j| < \tau_{s}$, we assume no cycle slip has occurred and set the ambiguity change to zero:
\begin{equation}
    \Delta \check{N}_{k-1,k}^{j} = 0, \quad \text{if } \text{Status}_{k-1}^j = \texttt{Hold} \land |\gamma_k^j| < \tau_{s}.
\end{equation}
Conversely, if the residual exceeds the threshold or the measurement is missing, the satellite is downgraded to \texttt{Drop} and temporarily excluded from anchor-based constraints.

\noindent \textbf{Recovery from Drop:} 
\texttt{Drop} measurements undergo a \textit{Tentative Tracking Mechanism}. To prevent unstable signals from corrupting the estimator, a satellite is upgraded back to \texttt{Hold} only if it maintains cycle-slip-free consistency over a sliding window of $M$ steps (experimentally set to $M=3$):
\begin{equation}
    \text{Status}_k^j = 
    \begin{cases} 
    \texttt{Hold}, & \text{if } \sum_{m=0}^{M-1} \mathbb{I}(|\gamma_{k-m}^j| < \tau_{s}) = M \\
    \texttt{Drop}, & \text{otherwise}
    \end{cases},
\end{equation}
where $\mathbb{I}(\cdot)$ is the indicator function. Once this stability criterion is met, signal lock is confirmed, and the measurements re-enter the estimator.

After the above update, we obtain a complete per-satellite cache for all satellites ever observed up to time step $k$:
\begin{equation}
\mathcal{M}_k \triangleq \left\{ \left(j,\ \text{Status}_k^j,\ \Delta \check{N}_k^j\right)\ \middle|\ j \in \mathcal{O}_{1:k} \right\},
\end{equation}
where $\text{Status}_k^j \in \{\texttt{Hold},\texttt{Drop}\}$ and $\Delta \check{N}_k^j$ is the maintained cumulative ambiguity increment w.r.t. its anchor epoch. This cache directly determines which satellites are eligible to form adjacent TDCP and epoch-to-anchor factors in the back-end optimization.

\subsection{Robust Initialization Framework}
To prevent initialization failures caused by transient satellite signal instability and cycle slips, we propose a cycle-slip-aware initialization framework (Fig.~\ref{fig:initialization_framework}) that operates in a coarse-to-fine stages:

\noindent \textbf{Coarse Alignment:} 
Without external priors, we first estimate the anchor position $^E\mathbf{p}_{\text{anc}}$ via single point positioning. We formulate alignment constraints incrementally using a compensated TDCP model \cite{fredaTimedifferencedCarrierPhases2015}. By relating the adjusted Doppler integral---corrected for satellite motion $\Delta D^j_k$ and geometry changes $\Delta g^j_k$---to the local relative position increment $\Delta^l\mathbf{p}_{k}$, we recover the coarse initial yaw $\psi_{0}$ and clock drift $\Delta \delta t_{k}$ over a window of size $N$:
\begin{equation}
\begin{split}
    \min_{\psi_0, \Delta \delta t_{1 \rightarrow N}} \sum_{k=1}^N \sum_{j \in \mathcal{S}_k} \Big\| & \lambda \Delta \tilde{\Phi}_{\mathcal{D},k}^{adj, j} + \mathbf{e}_k^{j^T} {}^E\mathbf{R}_l(\psi_0)\Delta^l\mathbf{p}_k \\
    & - c(\Delta \delta t_k - \Delta \delta t_{k-1}) \Big\|^2,
\end{split}
\end{equation}
where $\lambda \Delta \tilde{\Phi}^{adj,j}_{\mathcal{D},k} \approx \lambda \frac{\mathcal D^j_{k-1} + \mathcal D^j_k}{2} \Delta t - \Delta D^j_k + \Delta g^j_k$, and $\mathbf{e}_k^j$ is the unit line-of-sight vector. $\Delta \delta t_{1 \rightarrow N}$ represents the sequence of clock drift estimates over the initialization window. This ensures the system acquires a valid alignment before engaging precise phase constraints.

\noindent \textbf{Refine Alignment:}
As doppler measurements are susceptible to noise, especially when the signal-to-noise ratio (SNR) is low, we further refine the initial yaw estimate using TDCP measurements within the initialization window with the optimized $ \psi_{0} $.
\begin{align}
    r(\mathbf{\hat x}_k, \Delta{\Phi^{j}_k}) &= \| {}^E \mathbf p ^j_k -  (^E\mathbf{p}_{\text{anc}} + { }^E{\mathbf{R}_l(\hat \psi_{0})} {}\mathbf{\hat p}_k^l) \| \notag \\
    &\quad - \| {}^E \mathbf p ^j_{k-1} -  (^E\mathbf{p}_{\text{anc}} + { }^E{\mathbf{R}_l(\check \psi_{0})} {}\mathbf{\check p}_{k-1}^l) \| \notag \\
    &\quad - c (\Delta{\delta \hat t_{k}} - \Delta{\delta \check t_{k-1}}) - \Delta\Phi_{k}^{j}.
\end{align}
We optimize the yaw angle $\psi$ by minimizing the following cost function:
\begin{align}
	\min_{\psi} \sum_{k=1}^{N} \sum_{s \in \mathcal{S}_k} &\| r(\psi, \Delta{\Phi^{j}_k}) \|^2.
\end{align}

By solving this optimization problem, a high-fidelity initial alignment between the local frame and the global ENU frame is established. This refined yaw angle, along with the estimated clock drift, serves as a crucial initial seed for the subsequent sliding-window optimization.

\subsection{Factor-Graph-Based State Estimation}\label{sec:graph}
The factor graph consists of nodes representing the robot's states, and edges representing the measurements from the GNSS receiver. Followed are the details of each factor.
\begin{figure}[htbp]
	\centering
	\resizebox{\linewidth}{!}{ 
		\begin{tikzpicture}[node distance=2.5cm]
			\node[obs, fill=gray!30, minimum size=1.2cm] (x0) {\Large $\mathbf{x}_0$};
			\node[below=0.2cm of x0] {\large Anchor};
			\node[latent, right=of x0, minimum size=1.2cm] (x1) {\Large $\mathbf{x}_1$};
			\node[obs, fill=white, right=of x1, minimum size=1.2cm] (x2) {\Large $\mathbf{x}_2$};
			\node[right = 0.2cm of x2, minimum size=0.4cm] (dots) {\Huge $\dots$};
			\node[latent, fill=gray!30, right = 0.1cm of dots, minimum size=1.2cm,
				  label={[yshift=-0.2cm]below:{\scriptsize \color{red} \large New Sats $j$}}] (xi) {\Large $\mathbf{x}_{i}$};
			\node[below=0.7cm of xi] {\large Anchor};
			\node[right = 0.2cm of xi, minimum size=0.4cm] (dots2) {\Huge $\dots$};
			\node[latent, right=0.2cm of dots2, minimum size=1.2cm] (xkm1) {\Large $\mathbf{x}_{k-1}$};
			\node[latent, right=of xkm1, minimum size=1.2cm] (xk) {\Large $\mathbf{x}_k$};

			\draw[-, thick, blue] (x0) to[bend right=-45] node[midway, fill=blue, circle, inner sep=4pt] {} (x1);
			\draw[-, thick, blue] (x1) to[bend right=-45] node[midway, fill=blue, circle, inner sep=4pt] {} (x2);
			\draw[-, thick, blue] (xkm1) to[bend right=-45] node[midway, fill=blue, circle, inner sep=4pt] {} (xk);

			\draw[-, thick, orange] (x0) to[bend right=45] node[midway, fill=orange, circle, inner sep=4pt] {} (x1);
			\draw[-, thick, orange] (x1) to[bend right=45] node[midway, fill=orange, circle, inner sep=4pt] {} (x2);
			\draw[-, thick, orange] (xkm1) to[bend right=45] node[midway, fill=orange, circle, inner sep=4pt] {} (xk);

			\draw[-, thick, dashed, green!60!black] (x0) -- node[midway, fill=green!60!black, diamond, inner sep=2.8pt] {} (x1);
			\draw[-, thick, dashed, green!60!black] (x1) -- node[midway, fill=green!60!black, diamond, inner sep=2.8pt] {} (x2);
			\draw[-, thick, dashed, green!60!black] (xkm1) -- node[midway, fill=green!60!black, diamond, inner sep=2.8pt] {} (xk);

			\draw[-, very thick, red] (x2.north) to[bend right=45] node[midway, fill=red, rectangle, inner sep=4pt] {} (x0.north);
			\draw[-, very thick, red] (xi.north) to[bend right=45] node[midway, fill=red, rectangle, inner sep=4pt] {} (x0.north);
			\draw[-, very thick, red] (xkm1.north) to[bend right=45] node[midway, fill=red, rectangle, inner sep=4pt] {} (xi.north);
			\draw[-, very thick, red] (xk.north) to[bend right=45] node[midway, fill=red, rectangle, inner sep=4pt] {} (xi.north);	
			\draw[-, very thick, red] (xk.north) to[bend right=45] node[midway, fill=red, rectangle, inner sep=4pt] {} (x0.north);

			\matrix [
				draw, 
				below=0.2cm,
				anchor=north,
				column sep=0.7cm,
				ampersand replacement=\&
			] at (current bounding box.south) {
				\node [fill=blue, circle, inner sep=4pt, label=right:\large Position Increment] {}; \&
				\node [fill=orange, circle, inner sep=4pt, label=right:\large E2E TDCP] {}; \&
				\node [fill=green!60!black, diamond, inner sep=2.8pt, label=right:\large Random Walk Clock Bias] {}; \&
				\node [fill=red, rectangle, inner sep=4pt, label=right:\large E2A Anchor] {}; \\
			};
		\end{tikzpicture}
	}
	\caption{\textbf{Factor Graph of the Proposed System.} The red arcs denote the Anchor-based constraints ensuring low-drift estimation. Dashed green factors denote the random walk clock bias between adjacent steps. Dashed line means that there is a new found satellites at $\mathbf{x}_i$ state.}
	\label{fig:factor_graph}
\end{figure}
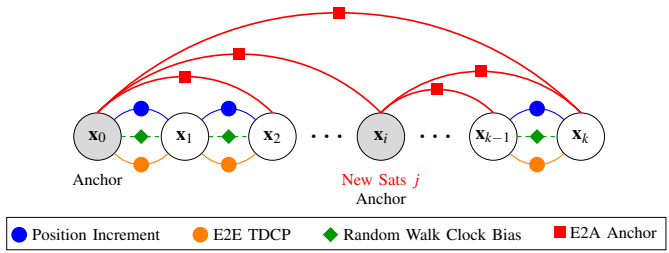

\noindent \textbf{Epoch-to-Anchor Difference Factor:}
To bound cumulative drift, satellites flagged \texttt{Hold} establish a global constraint linking the current time step $k$ to the virtual anchor step $i$. The residual $\mathbf{r}_{\mathcal{A}}$ is:
\begin{equation}
\begin{split}
    \mathbf{r}_{\mathcal{A}}(\boldsymbol{\chi}_k, \Delta{\Phi^{j}_{i,k}}) = & 
    \left\| {}^E \mathbf{p}^j_k - (^E\mathbf{p}_{i}^j + {}^E {\mathbf{R}}_l(\hat \psi_{k}) {}^l \mathbf{p}_k) \right\| \\
    & - \left\| {}^E \mathbf{p}_{i}^j - ^E\mathbf{p}_{\text{anc}}^j \right\| - c (\Delta{\delta t_{k}}\\
    & - \Delta \delta t_{i}) - \lambda \Delta\Phi_{{i, k}}^{j} + \lambda \Delta \check{N}_{{i,k}}^{j}
\end{split}
\label{eq:factor_anchor}
\end{equation}
where $\Delta{\Phi^{j}_{i,k}}$ is the phase difference between the anchor step $i$ and the current step $k$, $\Delta \check{N}_{{i, k}}^{j}$ is the recovered integer ambiguity and $\Delta{\delta t_{i, k}}$ is the clock bias variation. $\mathbf{p}_{i}^j$ is the satellite position at the anchor step $i$.

\noindent \textbf{Adjacent Difference Factor:} 
To constrain relative motion and ensure trajectory smoothness, we utilize TDCP measurements between consecutive steps $k-1$ and $k$. The residual $\mathbf{r}_{\mathcal{T}}$ is defined as:
\begin{equation}
\begin{split}
    \mathbf{r}_{\mathcal{T}}(\boldsymbol{\chi}_k, & \Delta{\Phi^{j}_{k}}) = 
    \left\| {}^E \mathbf{p}^j_k - (^E\mathbf{p}_{i}^j + {}^E {\mathbf{R}}_l(\hat \psi_{k}) {}^l \mathbf{p}_k) \right\| \\
    & - \left\| {}^E \mathbf{p}^j_{k-1} - (^E\mathbf{p}_{i}^j + {}^E {\mathbf{R}}_l(\hat \psi_{k-1}) {}^l \mathbf{p}_{k-1}) \right\| \\
    & - c (\Delta{\delta t_{k}} - \Delta{\delta t_{k-1}}) - \lambda \Delta\Phi_{k}^{j} + \lambda \Delta \check{N}_{k-1,k}^{j}.
\end{split}
\label{eq:factor_tdcp}
\end{equation}

Note that the ambiguity term $\Delta \check{N}_{k}^{j}$ is zero because of the cycle slip detection and maintenance strategy in section\ref{sec:cycleslipPar}. This factor provides a strong constraint on the relative motion in adjacent steps so that we can get a smooth trajectory.

\noindent \textbf{Position Increment Factor:}
Serving as a kinematic prior, this factor stabilizes frame alignment and maintains consistency during satellite ambiguity transitions (e.g., \texttt{Hold} to \texttt{Drop}). The residual $\mathbf{r}_{\mathcal{O}}$ between steps $k-1$ and $k$ is:
\begin{equation}
    \mathbf{r}_{\mathcal{O}}(\boldsymbol{\chi}_k, \Delta{{}^l \mathbf{\hat p}_{k}}) = 
    {}^e \mathbf{p}_k - {}^e \mathbf{p}_{k-1} - {}^e {\mathbf{R}}_l(\hat \psi_{k}) \Delta{{}^l \mathbf{\hat p}_{k}}.
    \label{eq:factor_odom}
\end{equation}

This constraint strictly aligns local odometry with the ENU frame, ensuring the accurate yaw ($\psi$) estimation prerequisite for our model-based cycle slip detection.

\noindent \textbf{Random Walk Clock Factor:}
Modeled as a random walk process~\cite{xuSinglestationSinglefrequencyGNSS2024}, this factor maintains temporal clock consistency across consecutive steps, particularly during satellite visibility transitions. The clock residual $\mathbf{r}_{\mathcal{C}}$ is defined as:
\begin{equation}
    \mathbf{r}_{\mathcal{C}}(\boldsymbol{\chi}_k) = 
    \Delta\delta{t}_k - \Delta\delta{t}_{k-1}.
    \label{eq:factor_clock}
\end{equation}
It's important to note that this factor is given a relatively low weight in the optimization, so it will only work when there is no other strong constraints (e.g., no \texttt{Hold} satellites) to prevent the estimator from diverging due to the lack of global constraints.

\noindent \textbf{Joint Optimization:}
The optimal state trajectory $\mathcal{X}^*$ is estimated by minimizing the joint non-linear cost function over the sliding window:
\begin{equation}
\begin{split}
    \mathcal{X}^* = & \mathop{\arg\min}_{\mathcal{X}} \Bigg( 
     \sum_{k} \sum_{j \in \mathcal{S}_{\mathcal{A}}} \rho_{\mathcal{A}} \left( \| \mathbf{r}_{\mathcal{A}} \|^2_{\mathbf{\Sigma}_{\mathcal{A}}} \right) + \\
    & \sum_{k} \sum_{j \in \mathcal{S}_{\mathcal{T}}} \rho_{\mathcal{T}} \left( \| \mathbf{r}_{\mathcal{T}} \|^2_{\mathbf{\Sigma}_{\mathcal{T}}} \right) + \sum_{k} \| \mathbf{r}_{\mathcal{O}} \|^2_{\mathbf{\Sigma}_{\mathcal{O}}} + \sum_{k} \| \mathbf{r}_{\mathcal{C}} \|^2_{\mathbf{\Sigma}_{\mathcal{C}}}
    \Bigg),
\end{split}
\label{eq:optimization}
\end{equation}
where $\| \cdot \|^2_{\mathbf{\Sigma}}$ denotes the squared Mahalanobis distance, and $\rho(\cdot)$ is the Huber robust kernel applied to suppress GNSS outliers and multipath effects. The proposed factor graph (Fig.~\ref{fig:factor_graph}) is constructed and optimized using GTSAM~\cite{dellaert2012factor}.

\section{Experiment Results and Discussions}

\begin{figure}[htbp]
	\centering
	\includegraphics[width=0.8\linewidth]{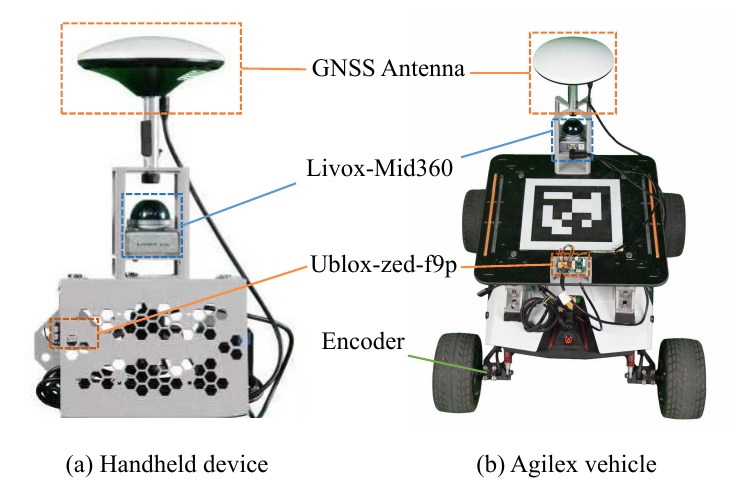}
	\caption{\textbf{Hardware Setup} for sequences collecting with LiDAR and encoder. (a) a handheld device. (b) a Agilex UGV platform.}\label{fig:set}
\end{figure}

\begin{figure}[htbp]
	\centering
	\includegraphics[width=1.0\linewidth]{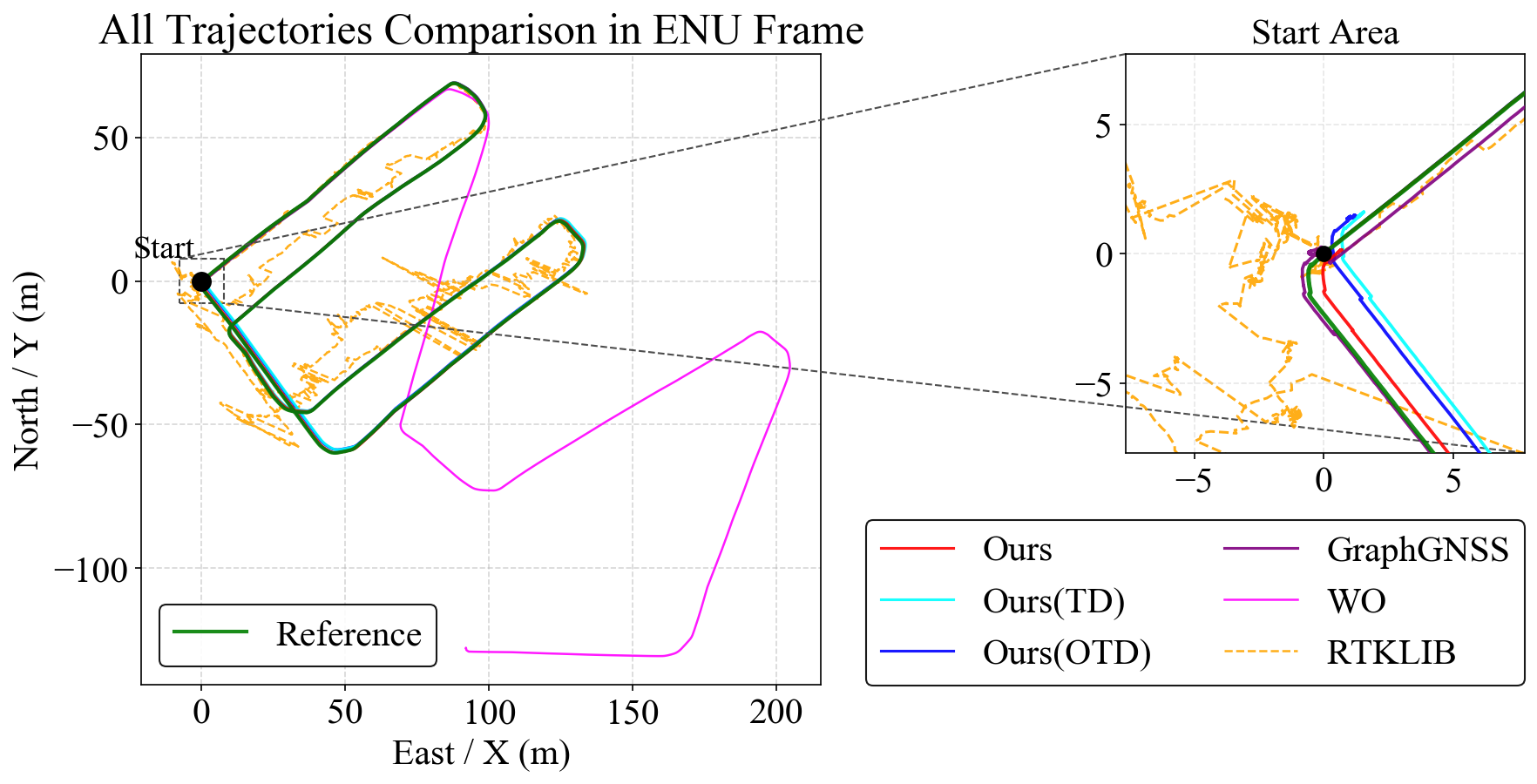}
	\caption{benchmark of the proposed method and baselines. The proposed method outperform in elimating drift. WO denotes wheel odometry.}\label{fig:bench_xy}
\end{figure}

We collected data in an expansive place (Fig.~\ref{fig:bench_xy}) and a challenging scenarios (Fig.~\ref{fig:pos}) with our experimental platform shown on Fig.~\ref{fig:set}.. During the test, a ublox-F9P GNSS receiver with WT-107 antenna is mounted to collect raw L1-frequency measurements at a frequency of 10 Hz. To establish ground truth, we utilizes stations (CORS) via the Internet. Besides, we run the proposed framework using an on-board Intel NUC featuring an i7-1260P processor. 
We evaluate our method in terms of positioning accuracy, stability to cycle slips, and the robustness in challenging scenarios.

\begin{table}[htbp]
    \centering
    \caption{Details of the Collected Sequences}
    \label{tab:dataset_details}
    \resizebox{\linewidth}{!}{ 
		\begin{tabular}{lccccc}
		\toprule
		\textbf{Seq.} & \textbf{Platform} & \textbf{Sensors} & \textbf{Dur. (s)} & \textbf{Len. (m)} & \textbf{noise/cycles} \\
		\midrule
		\textbf{Field} & Handheld & VIO + GNSS & 1508 & 2072 & 0.0129 \\
		\textbf{Deg1} & UAV & LiDAR + GNSS & 809 & 4380 & 0.0102 \\
		\textbf{Hand} & Handheld & LiDAR + GNSS & 82 & 102 & 0.0119 \\
		\textbf{Car}  & UGV & LiDAR + GNSS & 445 & 623 & 0.0271 \\
		\textbf{SCar}  & UGV & Encoder + GNSS & 500 & 739 & 0.0171 \\
		\textbf{FCar}  & UGV & Encoder + GNSS & 297 & 607 & 0.0137 \\
		\bottomrule
		\end{tabular}
    }
	\centering	
\end{table}

For the positioning evaluation, we compare the following methods:
\begin{itemize}
\item \textbf{API}: The Accumulative Position Increment method. 
\item \textbf{Odom}: The baseline SLAM algorithm.
\item \textbf{RTKLIB}: The RTKLIB SPP method.
\item \textbf{GraphGNSS}: The factor graph method~\cite{wenRobustGNSSPositioning2021} using pseudorange and doppler measurements.
\item \textbf{Ours(TD)}: Only adjacent differenced carriers are used.
\item \textbf{Ours(OTD)}: Reset ambiguity~\cite{lyuOptimalTimeDifferenceBased2021} when states turn \texttt{Drop}.
\item \textbf{Ours}: Try to recovery ambiguity in proposed factor graph framework. 
\end{itemize}

We evaluate the proposed framework on a diverse set of real-world sequences collected from \textbf{Handheld}, \textbf{UGV} and \textbf{UAV} platforms. Table~\ref{tab:dataset_details} summarizes the diverse sequences, which encompass a wide spectrum of sensor modalities including LiDAR, visual-inertial, and wheel encoders. Furthermore, the reported phase noise levels (measured in cycles) characterize the environmental complexity of each sequence, reflecting the varying degrees of signal degradation encountered during data collection. For encoder sequences, we use the odometry from wheel encoder directly. To further evaluate the scalability of the proposed method in complex, long-duration scenarios, we integrate the global anchor constraint into SLAM frameworks. Specifically, we loosely couple the VINS~\cite{qin2018vins} and Point-LIO~\cite{hePointLIORobustHighBandwidth2023} front-end with our epoch-to-anchor constraint using \textbf{Field} and \textbf{Deg1} sequences introduced in~\cite{cao2022gvins} and~\cite{he2025ligo}.



\begin{table}[htbp]
	\centering
	\caption{Positioning Performance Evaluation under Single-Frequency Configurations (Unit: m)}
	\resizebox{\linewidth}{!}{
	\begin{tabular}{l|cccccc}
	\toprule
	\multirow{2}{*}{\diagbox{\textbf{Method}}{\textbf{Sequences}}} & \multicolumn{4}{c}{\shortstack{API-fusion}} & \multicolumn{2}{c}{SLAM-Fusion} \\
	\cmidrule(lr){2-5} \cmidrule(lr){6-7}
	& \textbf{Hand} & \textbf{Car} & \textbf{SCar} & \textbf{FCar} & \textbf{Field} & \textbf{Deg1} \\
	\midrule
	\textbf{API/Odom} & 4.985 & 14.91 & 72.25 & 50.63 & 3.671 & 3.777 \\
	\textbf{RTKLIB} & 4.671 & 13.84 & 19.81 & 12.05 & 3.280 & 7.220 \\
	\textbf{GraphGNSS} & 0.501 & 1.219 & 1.360 & 0.388 & 2.625 & 3.679 \\
	\shortstack{\textbf{Ours} \textbf{(TD)}} & 0.103 & 0.206 & 1.157 & 0.451 & 2.106 & 3.667 \\
	\shortstack{\textbf{Ours} \textbf{(OTD)}} & 0.160 & 0.802 & 1.122 & 0.546 & 2.268 & 3.887 \\
	\textbf{Ours} & \textbf{0.085} & \textbf{0.198} & \textbf{0.525} & \textbf{0.270} & \textbf{1.335} & \textbf{3.392} \\
	\bottomrule
	\end{tabular}
	}
	\label{tab:rmse}
\end{table}

\subsection{Evaluation of Positioning Accuracy}

As shown in Table~\ref{tab:rmse}, the accumulative position increment suffers from significant unbounded drift over time. \textbf{RTKLIB} achieves very low positioning accuracy. \textbf{GraphGNSS} performs unstably: its accuracy is poor under weak signal conditions, while it remains relatively acceptable when the signal is strong. The \textbf{Ours(OTD)} method performs even worse. In contrast, the proposed method (\textbf{Ours}) effectively eliminates the cumulative drift by leveraging the epoch-to-anchor factors.
It's necessary to note that ionospheric delay can't be eliminated for single-frequency GNSS, which is the main source of our error.

\begin{table*}[!h]
\centering
	\caption{Cycle-Slip Detection Performance Across $C/N_0$ Bands.}
	\label{tab:slip_eval}
	\setlength{\tabcolsep}{6pt}
	\begin{tabular}{llccc cccc}
	\toprule
	$C/N_0$ (dB-Hz) & Method & MAE & RMSE & STD & Precision (\%) & Recall (\%) & F1-score
	(\%) & Recovery Acc. (\%) \\
	\midrule
	\multirow{3}{*}{$25 \leq C/N_0 < 30$}
	& Doppler  & 0.0575 & 0.0715 & 0.0690 & 20.00 & 100.00 & 33.33 & --- \\
	& Odometry & 0.0626 & 0.0767 & 0.0766 & 25.00 & 100.00 & 40.00 & --- \\
	& \textbf{Ours} & \textbf{0.0559} & \textbf{0.0692} & \textbf{0.0680} &
	\textbf{50.00} & \textbf{100.00} & \textbf{66.67} & --- \\
	\midrule
	\multirow{3}{*}{$30 \leq C/N_0 < 35$}
	& Doppler  & 0.0461 & 0.0601 & 0.0577 & 48.28 & 100.00 & 65.12 & 100.00 \\
	& Odometry & 0.0575 & 0.0729 & 0.0729 & 28.00 & 100.00 & 43.75 & 88.89 \\
	& \textbf{Ours} & \textbf{0.0424} & \textbf{0.0537} & \textbf{0.0531} &
	\textbf{66.67} & \textbf{100.00} & \textbf{80.00} & \textbf{100.00} \\
	\midrule
	\multirow{3}{*}{$35 \leq C/N_0 < 40$}
	& Doppler  & 0.0282 & 0.0366 & 0.0364 & 90.32 & 99.12 & 94.51 & 98.72 \\
	& Odometry & 0.0448 & 0.0575 & 0.0575 & 52.11 & 98.23 & 68.10 & 94.87 \\
	& \textbf{Ours} & \textbf{0.0259} & \textbf{0.0336} & \textbf{0.0336} &
	\textbf{96.58} & \textbf{100.00} & \textbf{98.26} & \textbf{100.00} \\
	\midrule
	\multirow{3}{*}{$40 \leq C/N_0 < 45$}
	& Doppler  & 0.0236 & 0.0307 & 0.0305 & 93.80 & 98.37 & 96.03 & 100.00 \\
	& Odometry & 0.0402 & 0.0519 & 0.0519 & 62.01 & 96.31 & 75.44 & 98.16 \\
	& \textbf{Ours} & \textbf{0.0199} & \textbf{0.0259} & \textbf{0.0258} &
	\textbf{100.00} & \textbf{99.59} & \textbf{99.79} & \textbf{100.00} \\
	\midrule
	\multirow{3}{*}{$C/N_0 > 45$}
	& Doppler  & 0.0224 & 0.0293 & 0.0291 & 93.75 & 100.00 & 96.77 & 100.00 \\
	& Odometry & 0.0382 & 0.0493 & 0.0493 & 60.00 & 100.00 & 75.00 & 100.00 \\
	& \textbf{Ours} & \textbf{0.0139} & \textbf{0.0178} & \textbf{0.0178} &
	\textbf{100.00} & \textbf{100.00} & \textbf{100.00} & \textbf{100.00} \\
	\midrule
	\multirow{3}{*}{All}
	& Doppler  & 0.0258 & 0.0341 & 0.0340 & 88.89 & 98.71 & 93.54 & 99.61 \\
	& Odometry & 0.0422 & 0.0545 & 0.0545 & 56.04 & 97.16 & 71.08 & 96.91 \\
	& \textbf{Ours} & \textbf{0.0223} & \textbf{0.0297} & \textbf{0.0297} &
	\textbf{96.98} & \textbf{99.74} & \textbf{98.34} & \textbf{100.00} \\
	\bottomrule
\end{tabular}
\label{tab:cycleslip}
\end{table*}

Table~\ref{tab:rmse} further demonstrates the effectiveness of the proposed method in improving the accuracy of SLAM loose fusion. By incorporating the global anchor constraints, our method significantly reduces the trajectory drift compared to the odometry-only and common TDCP-fusion methods, achieving a obvious improvement in positioning accuracy.


\subsection{Evaluation of Cycle Slip Detection and Recovery Method}
We assess cycle-slip detection with a injection framework on \texttt{FCar}. A slip is
injected by adding a persistent offset to a satellite's carrier phase from the
injection epoch onward. Injection magnitudes range from $0.2$ to $10$~cycles with random
sign. Each method declares a slip when the residual between the observed time-differenced carrier and
its predicted value exceeds $0.25$~cycle. For the small slips like $0.2$ and $0.5$, we just inject them once and check whether they are detected. The \emph{Doppler} baseline predicts from averaged Doppler
observations\cite{zhaoHighrateDoppleraidedCycle2020}, the \emph{Odometry} baseline from the
API-derived relative motion with the previous epoch's clock-bias change~\cite{xuSinglestationSinglefrequencyGNSS2024}.

As Table~\ref{tab:slip_eval} shows, our slip detector is the most accurate in
every CN/0 bin and every metric. Overall it lowers MAE by $13.6\%$ over the
Doppler baseline, raises F1 by $4.8$~pp, and reaches $100\%$ recovery accuracy,
the residual gain grows to $\sim38\%$ at the strongest signals where the TDCP
component dominates, while in the weakest band the Doppler-weighted blend
preserves robustness and roughly doubles F1 through fewer false alarms. The
\emph{Odometry} baseline, relying on the previous-epoch clock estimate, is consistently
weakest, confirming that the current-epoch clock refinement and the
variance-reduced fusion drive the improvement.

\begin{figure}[htbp]
	\centering
	\includegraphics[width=0.85\linewidth]{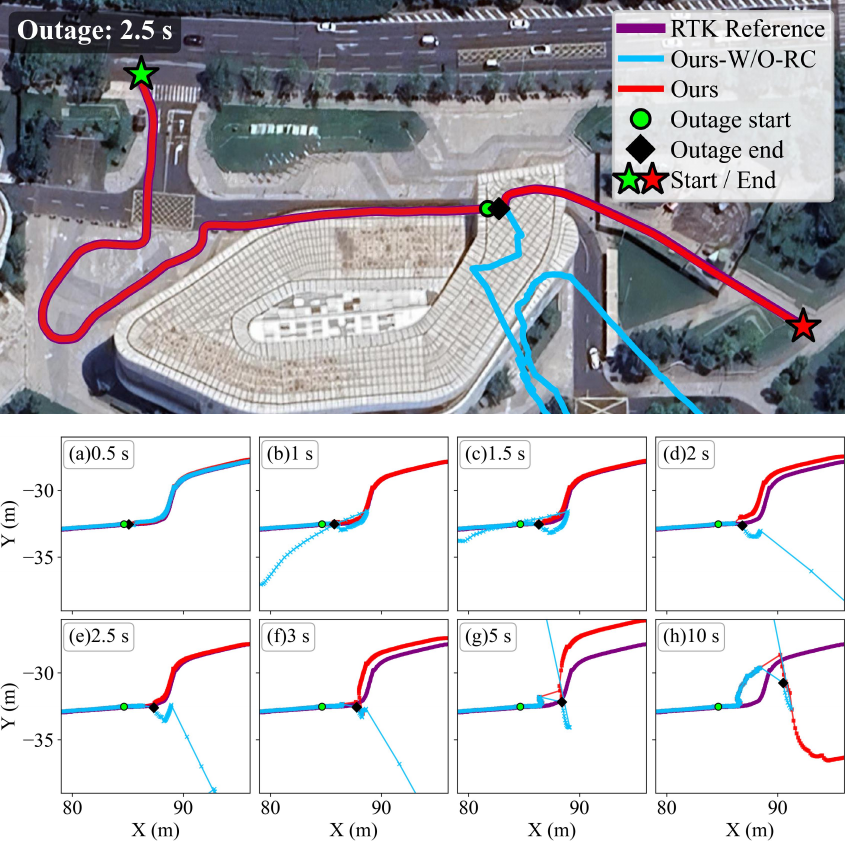}
	\caption{\textbf{Positioning Trajectories.} Ablation of the receiver-clock random-walk (RC) TDCP outages. (a)-(h) Close-ups around the outage region for increasing outage durations (0.5, 1, 1.5, 2, 2.5, 3, 5, 10 s).}\label{fig:pos}
\end{figure}

\begin{table}[htbp]
\centering
\caption{Maximum position drift (m) vs.\ TDCP outage duration.}
\setlength{\tabcolsep}{4pt}
\begin{tabular}{l|c|c|c|c|c|c|c|c}
Outage (s) & 0.5 & 1 & 1.5 & 2 & 2.5 & 3 & 5 & 10 \\
\hline
Ours        & 0.38 & 0.41 & 0.57 & 1.14 & 1.06 & 3.62 & 5.55 & 30.4 \\
Ours-W/O-RC & 0.40 & 74.1 & 73.3 & 169.8 & 154.2 & 325.1 & 8114 & 3295 \\
\end{tabular}
\label{tab_rc}
\end{table}

\subsection{Robustness Analysis}
We evaluate robustness under GNSS signal outages. As discussed in
Section~\ref{sec:graph}, the random-walk clock (RC) factor maintains the
observability of the receiver-clock state when no valid satellite
observation is available. To verify its effectiveness, we ablate the RC
factor while retaining all other factors, and compare against the full
system over outages of increasing duration.

As shown in Fig.~\ref{fig:pos}, without the RC factor the clock state
becomes unobservable once satellites are lost, and the clock error is
absorbed into the position estimate, causing the trajectory
(Ours-W/O-RC, blue) to diverge within roughly one second. In contrast,
the full system (Ours, red) maintains clock continuity and stays
consistent with the RTK reference. Table~\ref{tab_rc} quantifies
the maximum position drift: removing the RC factor yields tens of meters
of drift after one second. With the RC factor, the drift stays at the
decimeter-to-meter level for short outages ($\leq$2.5\,s). As a continuity-preserving prior, it bounds
short-term drift and lets the estimator recover once satellites reappear,
though it does not arrest drift indefinitely ($\geq$5\,s). This matches
the short, intermittent occlusions of urban canyons or foliage, confirming
the RC factor is essential under degraded satellite visibility.
\section{Conclusion}

In this paper, we present a novel state estimation framework that achieves high-precision localization by enpowering a low-cost single-frequency GNSS observations with relative observations. By introducing epoch-to-anchor constraints, we establish a virtual reference anchored to the initial epoch, allowing the use of single-frequency carrier phase measurements for low-drift localization. Our method demonstrates significant performance gains across diverse platforms. Furthermore, our cycle-slip detection and recovery strategy proves robust in challenging urban canyon environments by leveraging multi-modal kinematic priors. The proposed framework achieves reliable decimeter-level accuracy, providing a highly cost-effective and autonomous solution for real-time applications on resource-constrained systems.

\addtolength{\textheight}{-12cm}   








\bibliographystyle{IEEEtran}
\bibliography{root,gslam}

\end{document}